\documentclass[letterpaper, 10 pt, conference]{ieeeconf}  
\usepackage[backend=biber, style=numeric, sorting=none]{biblatex}
\usepackage{graphicx}
\usepackage{subcaption}
\usepackage{color,soul}
\usepackage{amsmath}
\usepackage{amssymb}
\usepackage{algorithm}
\usepackage{algorithmicx}
\usepackage{setspace}
\usepackage{cleveref}
\usepackage{tikz}

\newcommand\copyrighttext{%
  \footnotesize \textcopyright~2025 IEEE. Personal use of this material is permitted. Permission from IEEE must be obtained for all other uses, in any current or future media,
including reprinting/republishing this material for advertising or promotional purposes, creating new collective works, for resale or redistribution to servers
or lists, or reuse of any copyrighted component of this work in other works.}
\newcommand\copyrightnotice{%
    \begin{tikzpicture}[remember picture,overlay]
    \node[anchor=south,yshift=10pt] at (current page.south) {\fbox{\parbox{\dimexpr\textwidth-\fboxsep-\fboxrule\relax}{\copyrighttext}}};
    \end{tikzpicture}%
}

\crefname{equation}{Eq.}{Eqs.}
\crefname{figure}{Fig.}{Figs.}
\crefname{table}{Table}{Tables}

\AtEveryBibitem{\small
  \setlength{\itemsep}{0pt}
  \setlength{\parskip}{0pt}
  \setlength{\bibitemsep}{0pt plus 0.3ex}
  \setlength{\bibitemsep}{0pt}
  \setlength{\bibhang}{0pt}
  \setlength{\bibnamesep}{0pt}
  \setlength{\biblabelsep}{3pt}
}

\IEEEoverridecommandlockouts                              
\title{\LARGE \bf
Autonomous Vehicle Lateral Control Using Deep Reinforcement Learning with MPC-PID Demonstration}

\author{Chengdong Wu$^{1}$, Sven Kirchner$^{1}$, Nils Purschke$^{1}$, Alois C. Knoll$^{1}$
\thanks{\hspace*{-1em}$^{1}$ Technical University of Munich, Garching, Bayern, Germany}%
\thanks{\hrule}
\thanks{\hspace*{-1em}This research was funded by the Federal Ministry of Education and Research of Germany (BMBF) as part of the CeCaS project, FKZ: 16ME0800K.}
}

\begin{document}
\makeatletter
\let\@oldmaketitle\@maketitle
\renewcommand{\@maketitle}{\@oldmaketitle
}
\makeatother
\maketitle
\copyrightnotice
\thispagestyle{empty}
\pagestyle{empty}

\begin{abstract}
The controller is one of the most important modules in the autonomous driving pipeline, ensuring the vehicle reaches its desired position. In this work, a reinforcement learning based lateral control approach, despite the imperfections in the vehicle models due to measurement errors and simplifications, is presented. Our approach ensures comfortable, efficient, and robust control performance considering the interface between controlling and other modules. The controller consists of the conventional Model Predictive Control (MPC)-PID part as the basis and the demonstrator, and the Deep Reinforcement Learning (DRL) part which leverages the online information from the MPC-PID part. The controller’s performance is evaluated in CARLA using the ground truth of the waypoints as inputs. Experimental results demonstrate the effectiveness of the controller when vehicle information is incomplete, and the training of DRL can be stabilized with the demonstration part. These findings highlight the potential to reduce development and integration efforts for autonomous driving pipelines in the future.

\end{abstract}

\addtocounter{figure}{-1} 

\section{INTRODUCTION}
Autonomous driving has emerged as a prominent focus in the automotive industry. A vehicle with autonomous driving Level 5 is the ultimate purpose of many vehicle manufacturers \cite{SAEInternational.}. The controller transfers the planned path and trajectory to the real-time position of the vehicle. To simplify the problem, the controllers in longitudinal and lateral directions are designed separately, even though the trajectory tracking control is a joint control problem in the real world. Current motion control approaches include model-based and learning-based methods.

One of the widely used approaches for solving this task is the Model Predictive Control (MPC). It is used in industrial processes especially for constrained optimization problems \cite{Han.2018}. Designers benefit from the straightforward control strategy, but it nevertheless possesses intrinsic drawbacks. One of the problems is the tedious trial-and-error tuning task of the parameters, these may include the weights in the optimization goal \cite{Mehndiratta.2018} and the prediction horizon \cite{BOHN2021314}. In addition, the control performance of MPC is highly dependent on the vehicle model used. Simple models may fail to capture the vehicle's dynamic properties, while overly complex models increase computation time, compromising the controller's real-time performance. Furthermore, external disturbances and potential measurement errors in real-world scenarios can degrade the model's accuracy, thus affecting the effectiveness of the MPC controller. A straightforward approach is to add a PID controller that serves as the feedback part of the control loop \cite{mpcpid}. However, the tuning process is experience-based and lacks adaptability.

With the increasing driving data, the learning-based control approaches have also gained popularity, especially in complex driving scenarios. The learning approaches can be divided into two main classes: Imitation Learning (IL) and Reinforcement Learning (RL). IL is a means of learning and developing new skills from observing these skills performed by an expert, assuming that the expert behaves optimally \cite{il1}. Behavior Cloning, Direct Policy Learning, and Inverse Reinforcement Learning are the three main methods of IL. In RL, the agent is the learner and it can interact with the environment by choosing an action, and obtaining the reward accordingly. The training process in RL aims to develop a decision-making policy that maximizes the expected long-term reward. Nevertheless, the performance of IL is dependent on the quality of expert demonstration. It contributes to unpredictable performance of the controller when the application scenario is different from that where it is trained, while the sparse reward and compromise between exploration and exploitation \cite{Sutton.2020} also lead to difficulty in the training process of RL. Therefore, it is feasible to apply the IL part to the DRL training process, in which the IL is treated as the demonstration. The Reinforcement Learning from Demonstration (RLfD) addresses the exploration-exploitation problem and the demonstration guides the agent, especially at the beginning of the training phase. In addition, the performance of the agent in this case is not limited to what is learned from the expert. 


In this work, a novel concept of path-following control algorithm using DRL with MPC-PID as a control demonstrator is proposed. Experiments are conducted in different tracks designed in the CARLA \cite{carla} simulator. The MPC-PID controller is integrated into the training phase as the online information and can be decoupled for tests. Experimental results reveal that our DRL algorithm surpasses the performance of the demonstrating controller for lateral control tasks, especially when the vehicle information is not complete, and the training process can be accelerated with the online information. The key contributions of this paper include:

\begin{itemize}
\item An MPC controller with PID as feedback part is proposed as the baseline of the control task, which is used later as the demonstration for the DRL training.
\item A DRL controller using Deep Deterministic Policy Gradient (DDPG) is trained with the demonstration from MPC-PID controller.
\item Tests are conducted for various scenarios, and the performance of the demonstrating MPC-PID controller and DRL are compared.
\end{itemize}

\section{RELATED WORK}
The integration of learning-based methods with conventional control approaches has gained significant popularity over the past few years. These methods can be mainly divided into two categories:

\begin{itemize}
\item \textbf{Conventional control approach with optimization using DRL} 
\item \textbf{DRL-based control approach with demonstrations}
\end{itemize}

\subsection{Conventional Control approaches for vehicle lateral control}
Various model-based control methods can be applied to the vehicle lateral control problems. The Model Predictive Control (MPC) \cite{overview} is a widely used approach due to its ability to handle constraints and predict future states. It is a form of controller in which the control action is obtained by solving a finite horizon optimal control problem at each sampling instant online. The Receding Horizon principle indicates that the optimization problem is solved iteratively in each sampling instant, but only the first calculated control signal is applied. Other methods include Linear Quadratic Regulator (LQR) \cite{lqr} and Sliding Mode Control (SMC) \cite{smc}.

\subsection{Deep reinforcement learning}
Reinforcement Learning (RL) has gained significant popularity in recent years because of its adaptable application, especially for complicated control and decision-making tasks. The RL problem starts in the context of Markov Decision Process (MDP) \cite{Markov2006}, in which it is assumed that the state of the system at time $t+1$ only depends on the state and decision made at time $t$. RL differs from the normal
MDP because it can only be solved online by the interaction of the agent with the environment. The model-free RL learns policy or value function without knowledge or an explicit model of transitions and rewards, while in the model-based RL problem, an approximate model of the true MDP problem is learned based on the experience from interacting with the environment, and the solution process is similar to an offline MDP problem. Because of its simpler implementation and robustness to the inaccuracy of model information, model-free RL is applied in this paper. Commonly used algorithms include DQN \cite{dqn}, DDPG \cite{DDPG1,DDPG2}, PPO \cite{PPO}, and SAC \cite{sac}.

Gao et al. \cite{rlt2} implemented an Adaptive Cruise Control (ACC) by employing the idea from the policy iteration method in reinforcement learning. P{\'e}rez-Gil et al. \cite{rlt1} succeeded in applying both Deep Q-Network (DQN) and Deep Deterministic Policy Gradient (DDPG) to the control layer of autonomous driving vehicle, and the simulation is conducted in CARLA. The results reveal that both methods fulfill the control requirements, but DDPG performs better. Wang et al. \cite{rlt4} proposed an actor-critic style control framework for trajectory tracking of the autonomous vehicle under safety constraints. The expert strategy is leveraged and it takes over the control when safety constraints are not satisfied. Luo et al. \cite{rlt5} combined Stanley's trajectory tracking algorithm \cite{Stanley} and Proximal Policy Optimization-clipped (PPO2) for the path tracking problem with safety guarantee. The path-tracking controller designed by Yao and Ge \cite{rlt6} guides the autonomous vehicle using the Deep Deterministic Policy Gradient of the Double Critical Network (DCN-DDPG). Experimental results show that it offers good environment adaptability, speed adaptability, and tracking performance. The DDPG will be used as the baseline of the baseline of the RL algorithm applied in this paper.

\subsection{Conventional approach with DRL optimization}
Apart from the direct application of the DRL algorithms for the control tasks, researchers made several attempts to deploy DRL in the existing control methods for better performance and robustness. Ma et al. \cite{rlt3} proposed a fusion DRL and PID control for path tracking, in which the actor-critic framework is designed to optimize online the PID control parameters. In the context of MPC, one of the most important parameters influencing the performance of the controller is the weights of the objective function. The tuning process can be automated by reinforcement learning \cite{Mehndiratta.2018, BahaZarrouki.2020, objective} instead of manually by human experts. B{\o}hn et al. \cite{BOHN2021314} introduced RL to adapt the prediction horizon length for MPC to strike the balance between control performance and computational efficiency. Since the physical model used in MPC also influences the performance, there are attempts to switch the vehicle models for different application scenarios, which is called the Multiple MPC (MMPC) \cite{macombine, mmpc}. In addition to the deterministic switch of vehicle models, vehicle model parameters can also be learned and updated using DRL \cite{mpcmodel}.

\subsection{DRL-based approach with demonstrations}
Instead of being used for the optimization and improvement of conventional control methods, RL is also utilized directly for control tasks with expert demonstration. The Reinforcement Learning from Expert Demonstrations contains either prior knowledge or online knowledge \cite{Ramirez.2022}. To use the prior knowledge, Taylor et al. \cite{10.5555/2031678.2031705} for example propose a three-step method called Human Agent Transfer (HAT), so that the summarized control policy can be transferred to an RL agent. Liu et al. \cite{demoav1} proposed a control algorithm based on DDPG for the combined control of autonomous vehicles. The experience from the human expert is used in the $L_2$ norm to demonstrate the difference between expert action and the output action of the policy network. A modified Soft Actor-Critic algorithm with Imitation Learning is proposed by Liu et al. \cite{demoav3}. In their work, the learning goals of maximizing Q-functions and imitating the expert demonstration are combined. They add a Q-value regularization to combat the overfitting issue and boost the learning speed, which means the imitation loss is only referred to if the agent's performance does not outweigh the expert's.

The knowledge from the demonstration can also be used before the reinforcement learning process, which makes the solution a supervised learning problem. One of the most important contributions in this aspect is proposed by Hester et al. \cite{dqfd}. In their work, deep Q-learning is combined with IL, in which the control policy begins with a pre-training phase, where only the data of the demonstrations stored in the replay buffer are available and there is no interaction with the environment.

Our proposed method is similar to those introduced in \cite{demoav1} and \cite{demoav3}, with DDPG as the baseline for the RL algorithm due to its deterministic policy and continuous action space. Nevertheless, the choice between expert knowledge and current RL policy is adaptive in our work, and the expert demonstrations collected from MPC run parallelly with the RL agent during the training process.

\begin{figure*}[htb]
    \vspace{1em}
    \centering
    \includegraphics[width=0.96\textwidth]{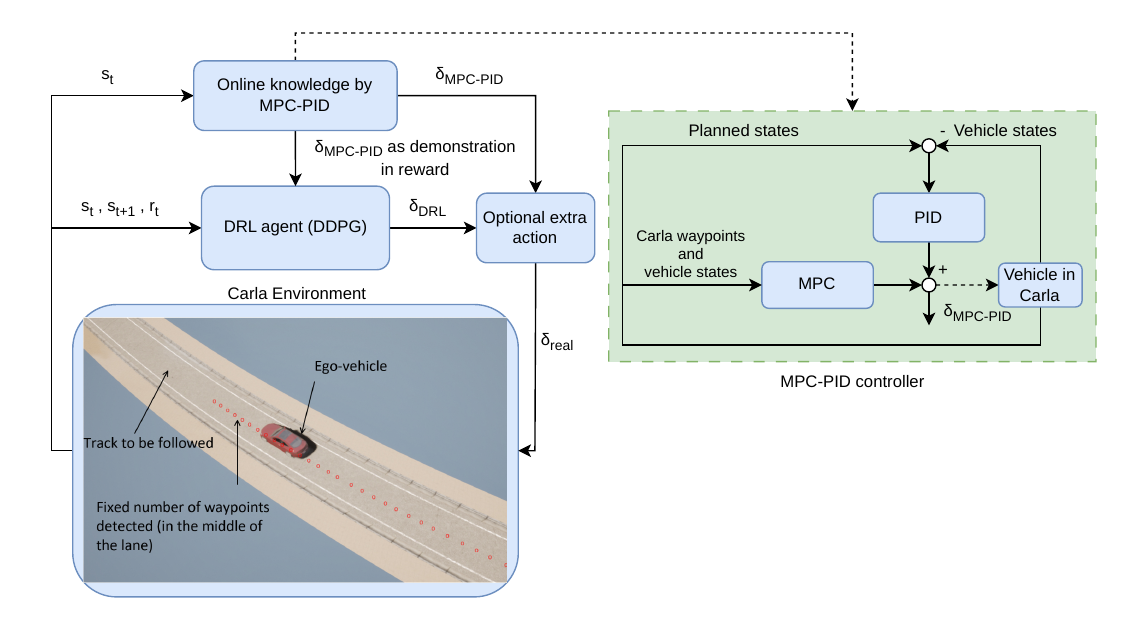}
    \caption{Overview of the proposed DRL controller with demonstration. The real steering command for the vehicle comes from both signals with a constant switch factor.}
    \label{fig:overview}
\end{figure*}

\section{Methodology}

\subsection{Overall learning framework}
The proposed approach can be divided into two parts, as shown in \cref{fig:overview}. The MPC-PID controller is designed as the baseline for the controller. A DRL controller is trained using DDPG with the online knowledge from the MPC-PID controller, which runs parallely with the DRL agent. The demonstrations are used in two aspects:

\begin{itemize}
\item Revised reward function
\item Extra action
\end{itemize}

In every simulation step, both MPC-PID and the DRL agent receive the vehicle state from the CARLA environment. The MPC-PID controller calculates the optimal control output directly with the receding horizon principle. The DRL agent additionally observes the reward resulting from the current action and the transition to the next state. The reward function includes not only the evaluation of path-following performance but also the IL part, which is used to mimic the demonstration from the MPC-PID controller. In addition, the steering action is decided not only by the current trained policy but also on the output of MPC-PID controller.

\subsection{MPC-PID controller}
The MPC controller is the base of our proposed framework, with the objective of minimizing the distance between the actual vehicle position and the predefined path, which is defined by the CARLA waypoints. To reduce the computational effort and therefore improve the control efficiency, a time-invariant continuous linear dynamic model is utilized in the calculation process of MPC. The state variables in the model include $X = [y \quad \beta \quad \psi \quad \dot \psi]$, which represents the lateral position, side-slip angle, yaw angle, and yaw rate respectively.

The optimization goal of the MPC is defined as:

\begin{equation}
\begin{split}
    \arg \min_{u} \int_{t_k}^{t_k+T_p} \, (\begin{bmatrix}
x - x_{ref} \\
y - y_{ref} \\
\end{bmatrix}^T
\textbf{Q}
\begin{bmatrix}
x - x_{ref} \\
y - y_{ref} \\
\end{bmatrix}
+
r \delta^2)
d\tau \\
+ 
\begin{bmatrix}
x(t_k + T_p) - x_{ref}(t_k + T_p) \\
y(t_k + T_p) - y_{ref}(t_k + T_p) \\
\end{bmatrix}^T
\textbf{P} \\
\begin{bmatrix}
x(t_k + T_p) - x_{ref}(t_k + T_p) \\
y(t_k + T_p) - y_{ref}(t_k + T_p) \\
\end{bmatrix}
\\
\text{subject to} \quad x(\tau) \in \mathbb{X} \\
\delta(\tau) \in \mathbb{U}
\end{split}
\label{eq:mpc}
\end{equation}

where the first term within integration denotes the stage cost at each step in the prediction horizon, and the r-term specifies the penalty factor for the input steering angle. The last term which lies out of the integration denotes the terminal cost at the end of the prediction horizon. $\textbf{P}$, $\textbf{Q}$, and $r$ represent weighting parameters. 

Nevertheless, the simplified vehicle model, vehicle parameters with limited accuracy, and external disturbances that cannot be modeled in most cases lead to unacceptable control performances. Solutions include more complex vehicle models or an online adaption of model parameters. In our framework, the problem is solved with the integration of PID feedback to reduce the computational effort. The final steering angle output is calculated by the combination of MPC and PID with hyperparameters. These are tuned together with the parameters in MPC.

\begin{equation}
    u(t) = c_{\mathrm{MPC}}u_{\mathrm{MPC}} + c_{\mathrm{PID}}u_{\mathrm{PID}}
\label{eq:add2}
\end{equation}

\subsection{Deep Reinforcement Learning with MPC-PID demonstration}
A DDPG agent is implemented in our work with the Stable Baselines3 \cite{sb3}. The lateral control task is modeled as an MDP task. Two parts are included in the observation of the DRL agent: (1) the ground truth of CARLA waypoints as the desired path, and (2) the current physical state of the vehicle. Detection and Planning modules are replaced by the ground truth of the path so that the control performance is not influenced by other modules in the autonomous driving pipeline.

In the autonomous driving control tasks the vehicles have continuous states, which contributes to an infinite amount of states for the RL agent. Therefore, state parameters of the ego-vehicle are used in our work with neural networks for value function estimation. Observations are flattened in the first step, followed by two fully connected layers with ReLU as the activation functions. 

The reward function aims to minimize the lateral deviation from the desired path and maximize the tangential velocity, and it is defined as:

\begin{equation}
        R = \sum_{t} |v_t cos\Phi_t| - |v_t sin\Phi_t| - |v_t||d_t|
    \label{eq:rewardfunction}
\end{equation}

where $v_t$ denotes the speed of the vehicle in scalar, $\Phi_t$ is the angle between the direction of the vehicle velocity and heading, and $d_t$ represents the shortest distance between vehicle and the planned path. Because DDPG outputs a concrete control signal, the inherent exploration characteristics does not exist. To let the ego-vehicle observe more physical states of itself and the simulation environment, the Ornstein-Uhlenbeck noise is integrated to the DRL training process. The output of the DRL network is either the steering angle itself or the derivation of the steering angle, with the latter chosen to limit steering fluctuations directly.

The pure DRL structure can provide persuasive control performance when the state space is explored thoroughly, and enough training time is given. It is trained on the computer with specifications shown in \cref{tab:hardware}.

\begin{table}[h]
\centering
\begin{tabular}{|l|l|}
\hline
\textbf{Name} & Simulation computer for DDPG agent training \\ \hline
\textbf{CPU} & Intel® Core™ i7-6850K 3.60GHz × 12\\ \hline
\textbf{GPU} & 2 x GTX 1080 16 GB VRAM\\ \hline
\textbf{RAM} & 32 GB DDR4-2133 RAM\\ \hline
\end{tabular}
\caption{Hardware specification of the device used for DDPG agent training}
\label{tab:hardware}
\end{table}

Early-stage DRL training can be safety-critical due to randomly initialized parameters leading to unstable actions. To address this, we integrate MPC-PID guidance into the training process, particularly at the beginning. Although MPC-PID may be suboptimal under varying velocities, the DRL agent evaluates its reliability using the proposed algorithm and decides whether to follow the demonstration or rely on simulation observations. Training runs synchronously with MPC-PID, ensuring the simulation advances only after all computations are completed at each time step.

Instead of pre-training, our approach uses MPC-PID as online demonstration to avoid extensive data collection. Given the same or fewer observations than the DRL agent, MPC-PID provides control suggestions at each step. The extra action method allows the agent to choose between its policy action and the MPC-PID suggestion based on a pre-defined probability.

Besides, the revised reward function can be used to achieve different control objectives. If the steering angle is utilized directly as the output of DRL agent, the change of the steering angle can be included in the reward function to minimize the fluctuation of output steering angle:

\begin{equation}
\begin{split}
    R_t = c_{\mathrm{track}} (|v_t cos\Phi_t| - |v_t sin\Phi_t| - |v_t||d_t|) \\
    - c_{\mathrm{change}} \frac{1}{2}(\delta_t - \delta_{t-1})^2
\label{eq:reward1}
\end{split}
\end{equation}

The online knowledge is used in the adapted reward function as well. In this case, the agent is guided by the suggestion from MPC-PID from the beginning of the training process:

\begin{equation}
\begin{split}
    R_t = c_{\mathrm{track}} (|v_t cos\Phi_t| - |v_t sin\Phi_t| - |v_t||d_t|) \\ 
    - c_{\mathrm{diff}} \frac{1}{2}(\delta_{t,demo} - \delta_{t})^2.
\label{eq:reward2}
\end{split}
\end{equation}

The reward function includes an additional term measuring the difference between the demonstrated and actual steering angles at each time step. Thus, the DRL agent learns both to follow the desired path and to imitate the MPC-PID behavior. Time-varying parameters balance learning from demonstration versus environment observations, adapting to different training phases. The reward function in \cref{eq:reward2} can be rewritten as:

\begin{equation}
    R_t = c_{\mathrm{track}}(t) r_{\mathrm{track}}(t)- c_{\mathrm{diff}}(t) r_{\mathrm{diff}}(t).
\label{eq:reward4}
\end{equation}

The coefficients are designed to change according to the values of their corresponding reward functions:

\begin{equation}
\begin{aligned}
c'_{\mathrm{track}}(t) &= S(r_{\mathrm{track}}(t))\\
c'_{\mathrm{diff}}(t) &= S(r_{\mathrm{diff}}(t))
\end{aligned}
\label{eq:coefficients}
\end{equation}

where the sigmoid function is defined as:

\begin{equation}
    S(x) = \frac{1}{1 + e^{-x}}.
\label{eq:sigmoid}
\end{equation}

The coefficients are then normalized. Additional parameters can also be added to emphasize one reward source over the other.

\begin{algorithm}[ht!]
    \caption{Pseudo code for Deep Deterministic Policy Gradient with MPC-PID Demonstration}
    \begin{algorithmic}%

    \State Input: initial policy parameters $\theta$, Q-function parameters $\phi$, empty replay buffer $\mathcal{D}$
    \State Set target parameters equal to main parameters $\theta_{targ} \leftarrow \theta$, $\phi_{targ} \leftarrow \phi$
    \State \textbf{Repeat}
    \State \indent Observe state $s$ and calculate the MPC-PID control output $a_{\mathrm{demo}}$
    \State \indent Observe state $s$ and select action $a = \text{clip}(\mu_{\theta}(s) + \epsilon, a_{\text{Low}}, a_{\text{High}})$, where $\epsilon \sim \mathcal{N}$
    \State \indent \textbf{if} $\mathrm{random}(0,1) < p_{\mathrm{action}}$
    \State \indent \indent Execute $a_{\mathrm{demo}}$ in the environment
    \State \indent \textbf{else}
    \State \indent \indent Execute $a$ in the environment
    \State \indent \textbf{end if}
    \State \indent Observe next state $s'$, reward $R(t)$ with two parts $r_{\mathrm{track}}$ and $r_{\mathrm{diff}}$, and done signal $d$ to indicate whether $s'$ is terminal
    \State \indent Store $(s, a, r, s', d)$ in replay buffer $\mathcal{D}$
    \State \indent If $s'$ is terminal, reset environment state
    \State \indent \textbf{if} it's time to update \textbf{then}
    \State \indent \indent \textbf{for} however many updates do
    \State \indent \indent \indent Randomly sample a batch of transitions, $B = \{(s, a, r, s', d)\}$ from $\mathcal{D}$
    \State \indent \indent \indent Compute targets
            \[
            y(r, s', d) = r + \gamma (1 - d) Q_{\phi_{\text{targ}}}(s', \mu_{\theta_{\text{targ}}}(s'))
            \]
            \State \indent \indent \indent Update Q-function using
            \[
            \nabla_{\phi} \frac{1}{|B|} \sum_{(s, a, r, s', d) \in B} \left( Q_{\phi}(s, a) - y(r, s', d) \right)^2
            \]
            \State \indent \indent \indent Update policy using
            \[
            \nabla_{\theta} \frac{1}{|B|} \sum_{s \in B} Q_{\phi}(s, \mu_{\theta}(s))
            \]
            \State \indent \indent \indent Update target networks with
            \[
            \phi_{\text{targ}} \leftarrow \rho \phi_{\text{targ}} + (1 - \rho) \phi
            \]
            \[
            \theta_{\text{targ}} \leftarrow \rho \theta_{\text{targ}} + (1 - \rho) \theta
            \]
   \State \indent \indent \textbf{end for}
   \State \indent \textbf{end if}
   \State \textbf{until} convergence
\end{algorithmic}%
\label{alg:ddpgwd}
\end{algorithm}%

\section{Experiments}
Both training and testing are conducted in CARLA using custom maps that combine straight segments with soft and sharp corners. The ground truth path is defined by lane-centered waypoints. Therefore, it is initially assumed that there are no external disturbances or detection errors. The vehicle is spawned at a given starting area, and it is expected to follow the ground truth of the lane. 

Two approaches are used in our work for the lateral control output. These include the direct method, which outputs the steering angle for control, and the indirect method, in which the derivation of steering angle is chosen as the lateral control output. In the second approach, the current steering angle is included additionally as the observation for the DRL agent. The steering angle for the next simulation step is calculated as:

\begin{equation}
    \delta_{t+1} = \delta_t + \dot\delta \Delta t.
\label{eq:derivatesteering}
\end{equation}

Because the MPC provides the steering angle as a demonstration, modification to the derivation of the steering angle is necessary for the demonstrated action in DRL training process.

\begin{equation}
    a'_{\mathrm{demo}} (t) = \frac{a_{\mathrm{demo}}(t) - a_{\mathrm{agent}}(t - 1)}{\Delta t}.
\label{eq:transfer}
\end{equation}

The corridor of track limit is designed to be 3 m, and that means the vehicle should be spawned again at the start point if the lateral error is larger than 1.5 m as a safety precaution. In the synchronous mode of CARLA, the simulation step is set to be fixed for 0.05 seconds. The scope of the steering angle is limited to [-1,1], representing a normalized and unitless value. If the derivation of the steering angle is chosen as the output, [-1.5708, 1.5708] $s^{-1}$ is the restriction.

The vehicle motion is decoupled into lateral and longitudinal directions. The throttle position can be held constant during the training process for simplicity. Despite this, we designed a basic PID controller for longitudinal direction so that the vehicle can simulate the real-world driving cycle in different velocities for adaptability. The throttle and brake inputs, which both lie within [0,1], are calculated according to the original control signal output of the longitudinal PID controller. The NEDC (New European Driving Cycle) is used as the training velocity profile with the first part of zero velocity removed.

It is desired to accelerate the training process at the beginning with a large learning rate and stabilize the learning outcome when approaching the optimal values. The cyclical learning rate \cite{learningrate} is used as the standard method. During the first half of the training process, the learning rate is kept constant to enable a fast approaching to the optimal parameters. In the second half of the training process, two constants are defined as the initial values for the cyclical learning rate decay: (1) The initial maximum learning rate $\eta_{max,0}$ is defined to be of the same value as the learning rate in the first half, and (2) the initial minimum learning rate $\eta_{min,0}$ is defined in our application as $\eta_{max,0} \cdot 10^{-1}$. The new learning rate bounds for each cycle are calculated as:

\begin{equation}
    \begin{aligned}
    \eta_{\min} = \eta_{\min,0} \cdot \lambda^{k_{\mathrm{cycle}}} \\
    \eta_{\max} = \eta_{\max,0} \cdot \lambda^{k_{\mathrm{cycle}}}
    \end{aligned}
\end{equation}

where $\lambda$ is an exponential decay factor and $k_{cycle}$ is the current cycle count. We choose 32 as the number of time steps in each cycle. The learning rate between the cycles is calculated directly using linear interpolation. The cyclical learning rate provides the neural network with the possibility to correct the value of learning rates during each training step, thus reducing the risk of inappropriate tuning.

After the training process is done with the demonstration, the MPC-PID controller can be detached from the control framework, making DRL the only source of the control signal. To verify the robustness of the controller, disturbances are added to the waypoints which are treated as possible detection errors. Random values within a specific range in both x and y directions are added to every waypoint. The test tracks include a perfect round circle for standardization and a customized racing track for adaptability and robustness.

\section{Results}
\subsection{Path-following results}
The vehicle is tested on different tracks except for those where it is tuned for MPC-PID and trained for DRL, with the waypoints defined in the middle of the track as well. The performance of the demonstrating MPC-PID controller and the DRL controller are tested and the results are compared with each other.

The test results indicate that the pure demonstrating MPC-PID controller is able to control the vehicle on the same track where it is tuned at a given velocity. Nevertheless, a different test rack and the changing velocity lead to unstable control performance, and the vehicle cannot drive through the whole track. An oversimplified vehicle model and parameter inaccuracies lead to noticeable trajectory deviations.

\begin{table}[h]
\centering
\begin{tabular}{|l|l|l|l|l|}
\hline
\textbf{Throttle position} & 0.4 & 0.6 & 0.8 & 1.0 \\ \hline
\textbf{ALE ($m$)} & 0.00283 & 0.0471 & 0.0853 & 0.104 \\ \hline
\textbf{AOE ($^\circ$)} & 0.922 & 2.12 & 3.87 & 5.01 \\ \hline
\textbf{SD of $\delta$} & 0.110 & 0.137 & 0.139 & 0.138 \\ \hline
\end{tabular}
\caption{Test results of DRL agent including Average Lateral Error (ALE), Average Orientation Error (AOE), and Standard Deviation (SD) in the circular circuit with $R=50m$ and different throttle positions. The steering angle is normalized into the range [-1,1].}
\label{tab:uniformresult}
\end{table}

\begin{figure}[htb]%
    \centering%
    %
    \includegraphics[width=40mm]{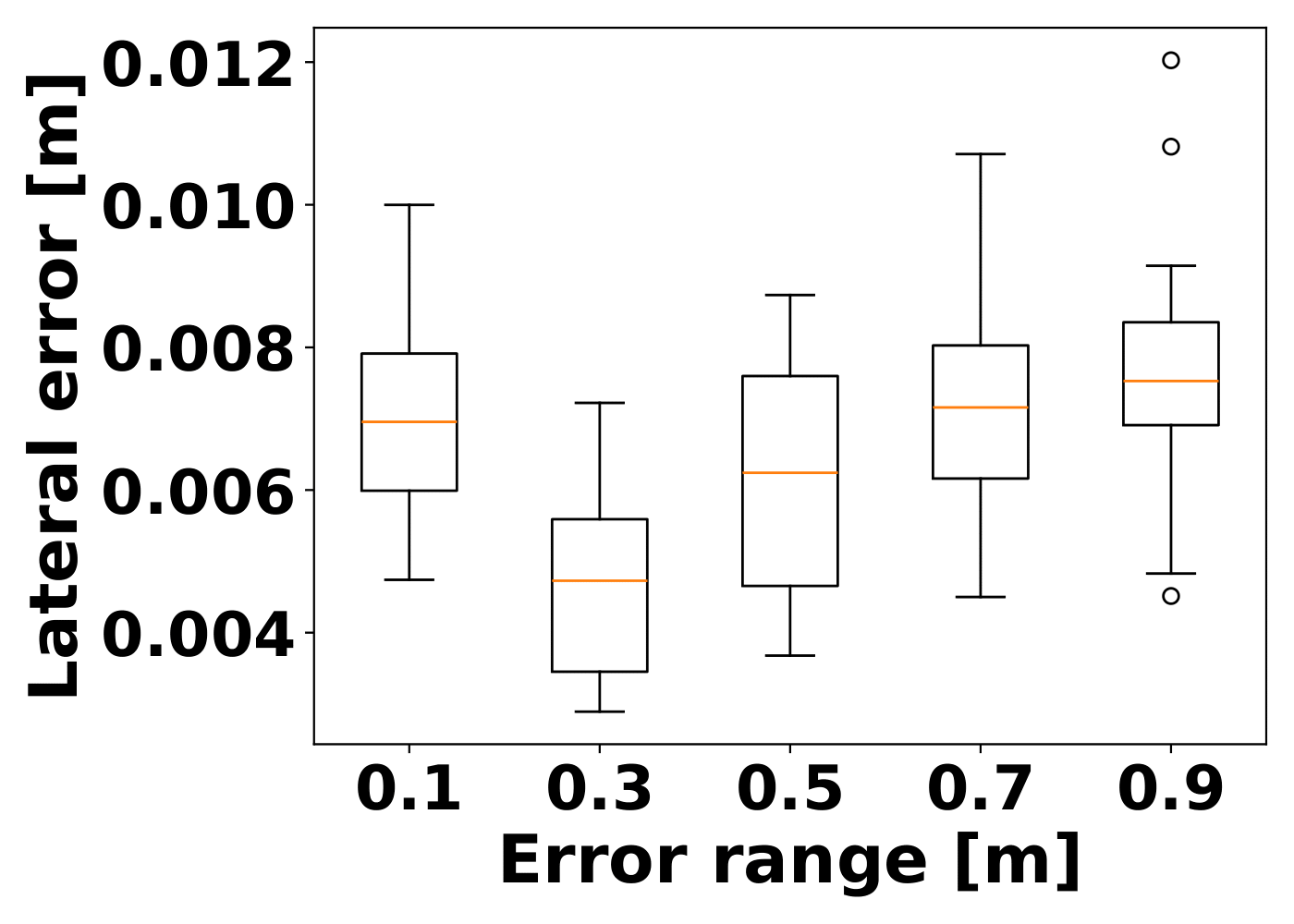}%
    \hspace*{1mm}%
    %
    \includegraphics[width=40mm]{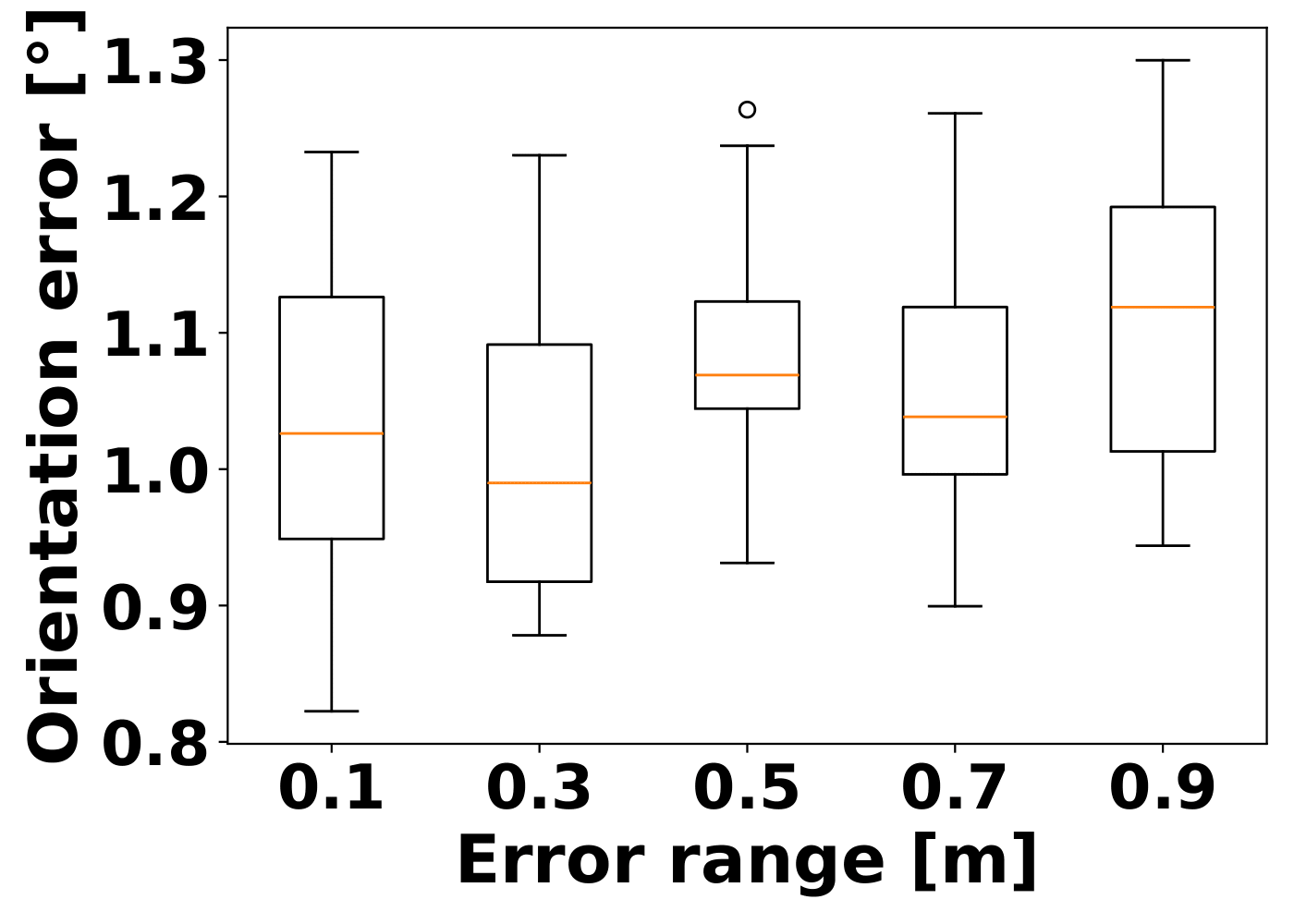}
    \caption{Test data for lateral error (left) and orientation error (right) with observation error in throttle position 0.5.}%
    \label{fig:robust}%
\end{figure}%

\begin{figure}[!htbp]
    \centering%
	\includegraphics[width=90mm]{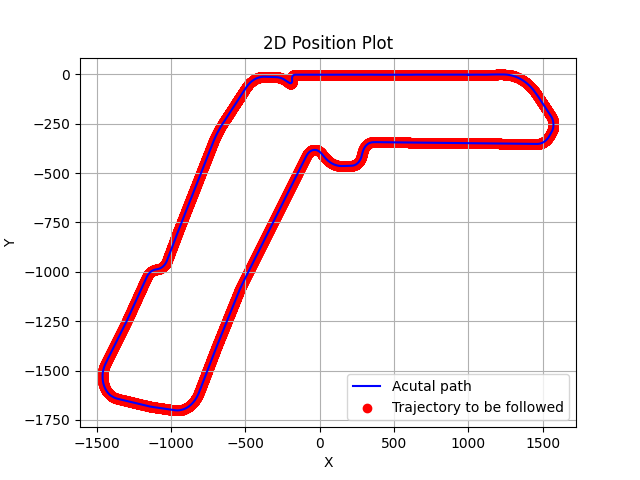}
	\caption{Relative position between ego-vehicle and the desired path with constant throttle position 0.4.}
	\label{fig:monza1}
\end{figure}

The trained DRL agent is first tested on uniform circular circuits with different velocities. The lateral errors and the orientation errors are shown in \cref{tab:uniformresult}. The results reflect that the increment of velocity results in the increment of path-following errors, but the vehicle is still able to follow the circular path with the full throttle. The waypoints are then added with random errors for the testing of robustness. Results in \cref{fig:robust} reflect that the vehicle can follow the trajectory with the existence of errors from other modules. The DRL agent is then tested on a racing track in the shape of Autodromo Nazionale Monza as shown in \cref{fig:monza1}. It can be seen from the result that the proposed DRL controller can guide the vehicle in an unfamiliar environment, thus verifying the adaptability. Because of the high number of waypoints observed by the vehicle, it performs a predictive behavior and in this way, reduces the long-term control cost.

The performance of both MPC and RL are tested in the circuit shown in \cref{fig:monza1} with the same throttle position 0.4 in synchronous mode. The running time of pure RL for the track is $145.28 s$, while the MPC-PID controller needs $664.23 s$ for the total calculation process around the track with a small prediction horizon. It turns out that RL can save about $78.13\%$ of the computational time when compared with the demonstrating controller. Additionally, the demonstrated MPC-PID controller yields acceptable tracking performance only at its tuned velocity, while it fails to maintain the vehicle within track limits at other speeds.

\subsection{Learning with demonstration results}
\begin{figure}[!htbp]
    \centering%
	\includegraphics[width=80mm]{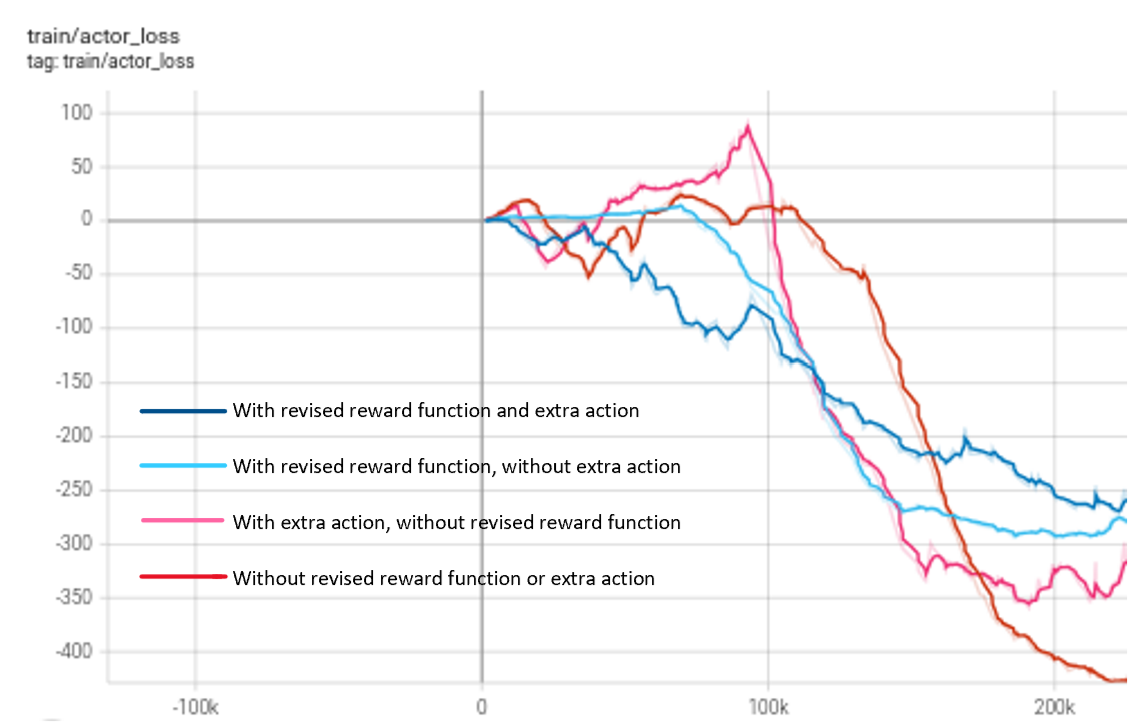}
	\caption{Actor losses for different approaches with and without demonstration.}
	\label{fig:loss}
\end{figure}

The proposed DRL training method with MPC-PID demonstration is evaluated in comparison with the conventional DRL methods. The training result is shown in \cref{fig:loss}. The actor loss in DDPG can be considered as the negative value of the action's expected Q-value, which means the objective of the DDPG agent is to minimize the actor loss. It is assumed that the hardware is not the bottleneck of the training speed. Hence the training speed only depends on the number of iteration periods. Only methods with the same reward function can be directly compared. The results indicate a smooth descending profile and little jump for the initial 100,000 training steps. From \cref{eq:reward2}, the second part of the revised reward function is always a non-positive value, indicating a smaller value of the resulting reward function than the original one. Because the first part of the reward function is considered as the expected reward to be compared, the actual expected reward of the agent with the revised reward function is higher than that shown in \cref{fig:loss}, representing a better performance of the cases with revised reward functions. In addition, when the methods of revised reward function and extra action are applied simultaneously, the DRL agent follows a smooth convergence.

\begin{figure}[htb]%
    \centering%
    %
    \includegraphics[width=40mm]{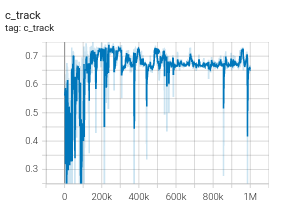}%
    \hspace*{1mm}%
    %
    \includegraphics[width=40mm]{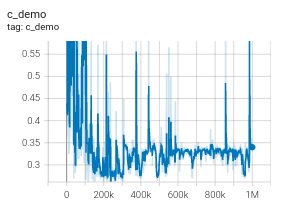}
    \caption{$c_{\mathrm{track}}$ (left) and $c_{\mathrm{demo}}$ (right) during the training process.}%
    \label{fig:coefficient}%
\end{figure}

To verify whether the adaptive decision process of the two reward sources works, the values of the time-variant coefficients are recorded as well during the training process, as shown in \cref{fig:coefficient}. The track reward possesses a larger proportion during the initial 100,000 training steps, and the proportions of these two reward sources are reversed for longer training time. That indicates more trust in imitation learning from demonstrating MPC-PID controller at the beginning and the gradual shift to the observation from the track when training proceeds. The profile of these time-variant coefficients shows that the proposed adaptive learning from demonstration method can work as demanded.

\section{Conclusions}
We propose a reinforcement learning framework with an online demonstration from the MPC-PID controller to combine the advantages of imitation learning and reinforcement learning. The MPC-PID demonstration is used for revising the reward function and taking extra action for exploration. A time-variant reward function is developed to adapt the learning objective, thus improving the convergence and performance of the learned DRL agent. The trained DRL controller is then tested on the uniform circular circuit and a racing track. The demonstration stabilizes and accelerates the training process of DRL agent, and the test result shows the improvement of performance and the adaptability of the trained DRL controller, compared with its demonstration.

With a series of waypoints forward as observations, the DRL agent can follow the path with the existence of the error from other modules in the autonomous driving pipeline, indicating robustness during the integration process. This also provides the ego-vehicle the predictive behavior by improving long-term expectations of the reward function. The proposed approach provides a general method for improving the performance of suboptimal model-based controllers.

\section{Future Work}
In future research work, there are areas that can be refined and improved. To start with, the separated control method in longitudinal and lateral directions can be replaced by a combined controller using reinforcement learning. Considering the influence of control performance between each direction, a combined controller improves the outcomes in both directions. Combined software concept of control can in this case reduce the computational effort and also result in a compact hardware layout. In addition, the source of the demonstration can also be improved in the future. In the real-world path and trajectory following scenarios, the understanding of drivers' intention in the surroundings and human-behavior imitation could also be one of the learning objectives. Therefore, human experts and, the outputs of Large Language Model (LLM) could also be potential demonstrators.

\begingroup
\setstretch{0.85} 
{\small
\printbibliography}
\endgroup


\end{document}